\pdfoutput=1

\documentclass[11pt]{article}

 \usepackage{acl}

\usepackage{times}
\usepackage{latexsym}

\usepackage{comment}

\usepackage[T1]{fontenc}

\usepackage[utf8]{inputenc}

\usepackage{microtype}

\usepackage{inconsolata}
\usepackage{booktabs}
\usepackage{multirow}
\usepackage{graphicx}
\newcommand{\todo}[1]{{\color{red} \textbf{#1}}}
\title{Comprehensive Evaluation for a Large Scale Knowledge Graph Question Answering Service}

\author{Saloni Potdar, Daniel Lee, Omar Attia, Varun Embar,  De Meng, \\   \textbf{Ramesh Balaji, Chloe Seivwright, Eric Choi, Mina H. Farid,} \\ \textbf{Yiwen Sun,  Yunyao Li} \\
        Apple}

\begin{document}
\maketitle
\begin{abstract}
Question answering systems for knowledge graph (KGQA), answer factoid questions based on the data in the knowledge graph. KGQA systems are complex because the system has to understand the relations and entities in the knowledge-seeking natural language queries and map them to structured queries against the KG to answer them. In this paper, we introduce Chronos, a comprehensive evaluation framework for KGQA at industry scale.  It is designed to evaluate such a multi-component system comprehensively, focusing on (1) end-to-end and component-level metrics, (2) scalable to diverse datasets \& (3) a scalable approach to measure the performance of the system prior to release. In this paper, we discuss the unique challenges associated with evaluating KGQA systems at industry scale, review the design of Chronos, and how it addresses these challenges. We will demonstrates how it provides a base for data-driven decisions and discuss the challenges of using it to measure and improve a real-world KGQA system.

\end{abstract}

\section{Introduction}

Knowledge graph question answering (KGQA) systems enable users to query structured data in the knowledge graph using unstructured, natural language queries \cite{zheng:vldb18}.
Industrial KGQA systems need to generate answers with low latency and often use KGs with billions of entities.
These are sophisticated systems consisting of several interacting components.
Evaluating the quality and reliability of KGQA systems is crucial for the end user to have a good experience \cite{tan2023evaluation, rodrigo2017study, radev2002evaluating, ong2009measurement}.
Moreover, this evaluation helps developers identify shortcomings of the existing system, and steps needed for improvement. 
This feedback loop is critical for the rapid development of the KGQA system.

KGQA system evaluation involves analyzing the performance of the various components that map the user question to a structured query and the accuracy and coverage of the data present in the knowledge graph.
This evaluation is challenging as the underlying KG is constantly evolving to capture new entities and events, and to ensure that the existing information is not stale.
Several KG evaluation techniques have been proposed in literature  \cite{gao2019efficient, ojha:emnlp17, paulheim2017knowledge}.
However, they do no evaluate the KG as a part of a larger KGQA system.  
In this work, we delve into the challenges of building a robust, reliable and efficient framework for evaluating a KGQA system.
Our framework, Chronos, is an amalgamation of automated testing techniques, human-in-the-loop studies, and domain-specific evaluation.  The major components of our evaluation framework, Chronos, is shown in Figure \ref{sys-eval-fig}. The data collection component considers various slices of datasets (e.g. general knowledge, temporal, geo-sensitive, etc.) to cover a plethora of usecases for open-doamin QA. We collect high quality component level gold-labels through carefully designed human grader task and replay these queries through KGQA system and collect comprehensive metrics. This evaluation framework helps us in deciding whether the new version of KGQA system is ready for launch, while also helping us in making prioritization decisions. The computed metrics are tracked and monitored on a dashboard which is updated periodically to assess any quality improvements or degradation over long periods of time. We hope our modular and comprehensive framework provides a basis for developing future KGQA evaluation systems. 

The main contributions of this paper are:
\vspace{-5pt}
\begin{itemize}
    \setlength{\itemsep}{-5pt}
    \item A modular and adaptable evaluation framework for enterprise KGQA systems. 
    \item A comprehensive failure ontology for single-hop KGQA and their measurement strategies. 
    \item A detailed case study and downstream understanding of the proposed framework.
\end{itemize}

\section{Preliminaries}
\label{prelim}
We briefly describe the system architecture of a typical KGQA system in terms of its components:

    \noindent \textbf{Entity Linking:} Entity linking identifies entity mentions present in the query and links them to the entities in the knowledge graph. The entity linking system consists of sub-components that detect mentions, retrieve candidates and rank the candidates based on the context.  
    
    \noindent \textbf{Relation classification:} This component identifies the relation in the user query.  
    
    \noindent \textbf{Structured Query Generation: } This component maps the user query to a structured machine-executable query. It uses the identified entities and the relation from the previous components to generate the query.  
    
    \noindent \textbf{Fact Retrieval:} This component executes the structured query on a KG index and returns an entity ID or a fact. The answer is to provide a verifiable and rich interactive experience to the end-user.

\section{Design Desiderata}

While evaluating real world KGQA systems, we need to consider various factors:

\noindent\textbf{1. Complexity:} As discussed above, KGQA systems are comprised of several components which are dependent on each other to generate the answer shown to the user. Failure of one component yields to an inexact answer due to this interdependence. For example, the type of the entity retrieved and the relation influences the mapping of the user query to structured query. In our framework, this is addressed by reporting the detailed metrics such as component level metrics alongside the end-to-end metrics.

\noindent\textbf{2. Generality:} KGQA systems need to answer queries that span various domains such as sports and music apart from general knowledge questions. Ongoing events and trending topics which usually pertain to time-sensitive usecases can reveal the system's capability to answer with up-to-date information, which can impact user trust. 
Additionally, inaccurate responses to queries related to high-profile entities can lead to sub-optimal results. A single evaluation dataset can not capture all the complexities, so we need to define a process to capture latest queries across various patterns and usecases in a repeatable way, while making sure that the framework is extensible to them.

\noindent\textbf{3. Repeatability:} 
Industry-scale KGQA components evolve continuously and independently as new features are added and existing features are improved over time. Several team members often work in parallel modifying different components such as relation classification, entity linking, KG query generation and fact retrieval systems, trying to make difficult trade-offs between accuracy and performance. For the specific task of KGQA, the freshness of information is crucial, so we need to make sure the underlying KG is constantly updated with new information. Continuous, end-to-end system evaluation 
can ensure that new changes do not lead to regressions. 



\begin{figure*}[t!]
\centering
\includegraphics[width=16cm]{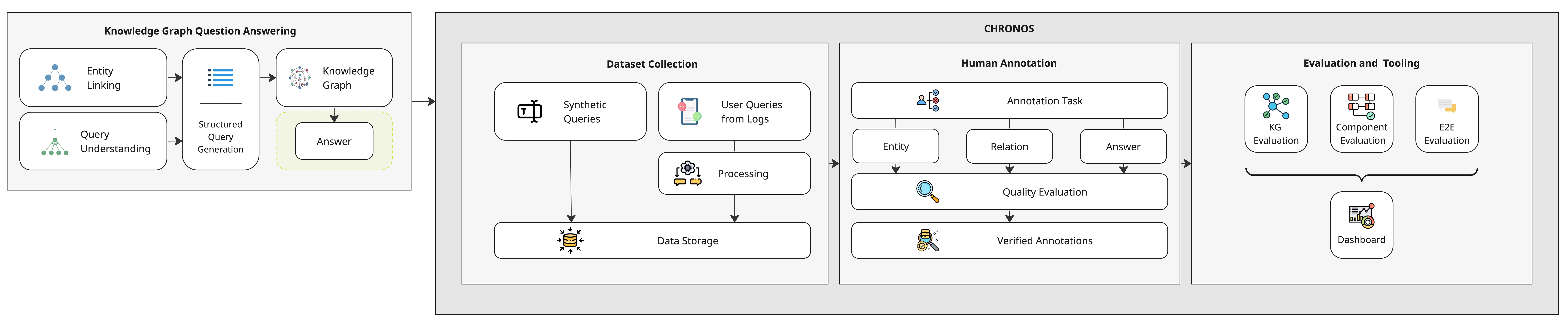} 
\caption{Chronos: Evaluation System Design for Knowledge Graph Question Answering}
\label{sys-eval-fig}
\end{figure*}

\section{Overview of Chronos}

In this section, we talk about the components of Chronos, where the design of the pipeline is informed by design considerations such as ease of debugging, independent development of components, comprehensive coverage of usecases, repeatability, and availability of metrics at various granularities. 

\subsection{Dataset Collection Component}
\label{sec:data_collection}
In research, the datasets for KGQA usually consist of a question and the answer such as MKQA \cite{Longpre2020MKQAAL}, ComplexQuestions, \cite{Bordes2015LargescaleSQ}, SimpleQuestions \cite{Bao2016ConstraintBasedQA} and Mintaka \cite{Sen2022MintakaAC}. More often than not, these datasets do not consist of component level gold-labels and are not comprehensive for industry-scale KGQA system. In our framework, the evaluation datasets are collected from optted-in \& privatized usage logs and through synthetic data generation to comprehensively cover all usecases, domains and data characteristics. 
We create evaluation datasets by creating a traffic weighted sample of logs. We also sample unanswerable queries and escalations which are raised by users of the KGQA service and use them to evaluate the system. 
In any production KGQA system the users inadvertently adapt to asking the system questions that it can answer. Thus, any evaluation dataset constructed solely from opted-in \& privatized user logs will be insufficient to comprehensively evaluate the system. 
Hence we augment this dataset with synthetically generated query sets like in  \cite{ravichander-etal-2021-noiseqa}. This can help in improving coverage over usecases deemed important for KGQA product by creating challenging query subsets by add several variations of the user-generated queries: 
\vspace{-6pt}
\begin{itemize}
    \item We replace the primary entity in the unanswerable query, by searching the graph for entities of the same ontology type 
    \vspace{-7pt}
    \item We paraphrase the query using paraphrase generation approaches \cite{zhou-bhat-2021-paraphrase} 
\end{itemize}

In addition to this, a simple classification strategy is used across all datasets (mined or generated) for query categorization. We use the relations and entity types in a query to determine which domain the query belongs to. This helps us track the quality of questions across domains such as people, albums and songs, sports, movie and TV shows, events, and general Q\&A.


\subsection{Human Annotation Process}
\label{sec:human_annotation} 

\begin{figure*}[t!]
\centering
\includegraphics[width=14.5cm]{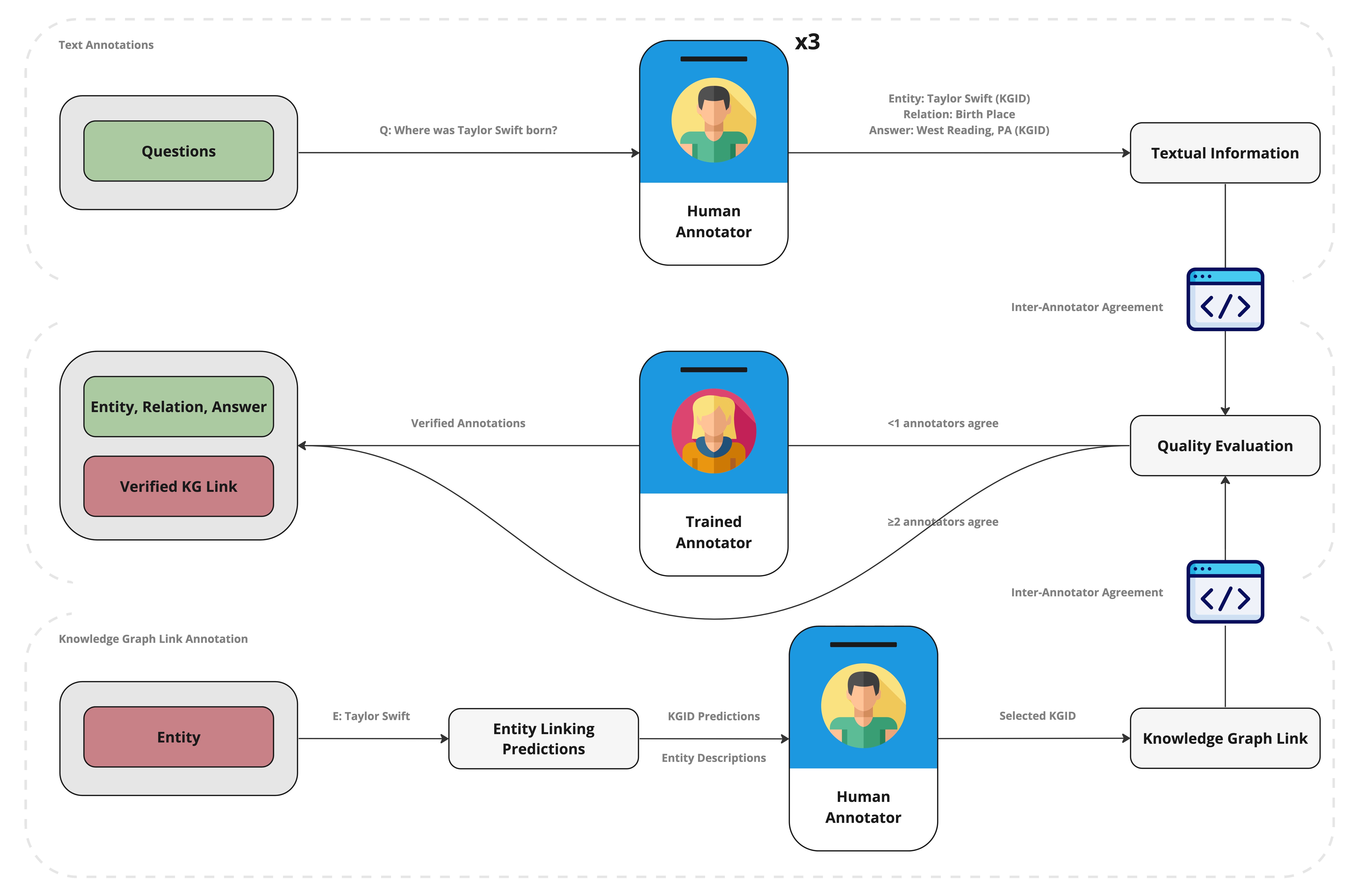} 
\caption{Human annotation process for Knowledge Graph Question Answering in Chronos}
\label{human-annotation}
\end{figure*}
To evaluate the accuracy of the answers to user queries, human graders annotate each query in the dataset. The UI design and guideline have been validated through pilot studies to ensure high quality gold labels. 
The overarching steps used in the grading process are as follows:


\noindent \textbf{Query Properties} 
Human graders are asked to define if an utterance is knowledge-seeking or fact-seeking by researching the knowledge base and the web pages. This allows us to filter out the KGQA queries from non-factoid queries (opinion-seeking, comparison, etc.). We also ask the graders if the query is geo-sensitive, time-sensitive, ambiguous, complex queries (similar to the classification in \cite{Sen2022MintakaAC}), etc. This helps us in breaking down the performance of the KGQA system on different data slices. 


\noindent \textbf{Entity \& Relation Annotation.} This helps us capture the entity span, entity ID from thr knowledge graph and relation within the query.


\noindent \textbf{Answer Annotation.}  The answer is captured and the entity ID is linked by researching the knowledge graph. The answer type is categorized into categories \cite{Longpre2020MKQAAL} Entity, Long Answer, Unanswerable, Date, Number, Number with Unit, Short Phrase and Binary. 

\noindent\textbf{Knowledge Graph Quality}: The quality of KG in terms of correctness of facts, completeness and freshness of entities and facts is crucial to the overall accuracy of the KGQA system. We use the query sets and facts derived from the human annotation process to continuously inform the quality and ensure high coverage of KG. 




Human annotation is critical for the evaluation of KG and KGQA systems, given its role in building datasets for validation and testing. However, it can be a source of inconsistency due to the inherent subjectivity and variability in human grading. This emphasizes the need for the objective and quantitative evaluation of the quality of data produced by human annotators and the performance of the annotators themselves. 
Two methods are utilized to measure the quality of human annotations: 

\noindent \textit{Inter-Annotator Agreement} We use Krippendorff's Alpha and Cohen's Kappa to measure the quality of annotated data. 

\noindent \textit{Human Annotator Tests} Qualification exams and consistency checks with a high qualification threshold is set to maintain a quality group of human graders.

\subsection{KGQA System Predictions Scraper}
\label{sec:sys_pred}
We designed a component to provide generalized interface that takes a query from datasets, replays query understanding via service API to retrieve corresponding entities and relations as mentioned in the query and answer. These predictions helps in assessing the quality of the system against gold labels collected from human annotation process as described in Section \ref{sec:human_annotation}.

\subsection{Evaluation and Tooling}
\label{sec:eval_metrics}
\subsubsection{Metrics}
The KGQA system is evaluated to produce system level metrics as well as component level metrics.
The system level metrics help us assess the overall correctness of the KGQA system, giving us an insight into which component led to the failure of the overall query. The end-to-end (E2E) system metrics are: 
    
    \noindent \textbf{E2E coverage}: total number of queries with relation and entity prediction over total number of queries.
    
    \noindent \textbf{E2E Precision}: total number of queries with correct relation and entity predictions over that of queries with relation and entity predictions.

The component metrics are computed in two ways: (1) view the component individually assuming that all previous component outputs are 100\% accurate to determine the headroom for the next component; (2) view the component conditioning on the correct predictions from previous components in order to access the actual performance in the cascading QA system.

While the end-to-end system level evaluation provides a holistic view of the quality in KGQA service, it does not answer specific questions regarding quality of data in the graph, like which domain has higher accurate facts, which relations and entity types have more missing facts. To be able to answer those questions, component-level evaluation of knowledge graph quality is instrumented with human-in-the-loop assessment.
These metrics are described in further detail in Section \ref{sec:eval_and_tooling}.

\subsubsection{Loss Bucketization of Errors}
\label{sec:loss_buckets}
Automated loss bucket analysis to highlight problems in various components of the QA pipeline which would facilitate the problems to reach the appropriate stakeholder for quick resolution. Loss bucketization is an automated root cause analysis service, capable of identifying errors on the relation, entity and fact retrieval. The buckets are largely classified into Query Understanding errors (QUE) and Knowledge Graph errors (KGE). They are further broken down into categories to highlight specific problems in the KGQA pipeline.

    \noindent \textbf{[QUE] Unsupported relation:} Predicted relation does not match with supported relation list or, the predicted relation is empty and gold relation for the query does not exist
    
    \noindent \textbf{[QUE] Relation Prediction Error:} Predicted relation by KGQA does not match gold relation
    
    \noindent \textbf{[QUE] Entity Prediction Error:} Predicted entity match does not match dominant gold label entity
    
    \noindent \textbf{[KGE] Missing Entity:} Gold entity not in KG
    
    \noindent \textbf{[KGE] Execution Error:} KG query failure
    
    \noindent \textbf{[KGE] Incorrect Fact:} Retrieved fact from KG does not match gold label fact
    
    \noindent \textbf{[KGE] Missing Fact:} Predicted relation and entity match gold labels but relevant fact absent in KG

\subsubsection{Dashboard}
The results from the KGQA system are tracked and monitored on a dashboard which is updated periodically. We also track and ensure the quality of the knowledge graph in addition to its ability to support the downstream tasks like KGQA on various slices of the datasets. The dashboard supports decision making for several partners in an industry setting.
The metrics tracked and how it supports decision-making is explained with an example in Section \ref{sec:eval_and_tooling}.

\section{Case Study}
\label{sec:eval_and_tooling}
In this case study, we use Chronos to assess multiple versions of the KGQA system and to assess Knowledge Graph quality. The evaluation was conducted on $\sim20000$ queries for two internal systems, namely System 1 and System 2. The data set covers over $\sim200$ unique relations and $\sim12000$ unique entities. 

The E2E coverage is close for both systems, System-1 out-performing System-2 marginally. The E2E precision and coverage for System-1 and System-2 is shown in Table \ref{tab:e2e-metrics}. The KGQA System-1 is further to yield component-level metrics. The relation prediction, entity linking and answer prediction component-wise results are shown in Table {\ref{tab:comp-metrics}}. This shows us that on the challenging data slice Dataset-3 the performance of all component drops, with Answer Prediction component struggling the most. This warrants further investigation on KG query computation and KG coverage. 

\begin{table}[]
\centering
\resizebox{\columnwidth}{!}{%
\begin{tabular}{@{}lllll@{}}
\toprule
\multicolumn{1}{c}{\multirow{2}{*}{}} & \multicolumn{2}{c}{\textbf{System 1}} & \multicolumn{2}{c}{\textbf{System 2}} \\ \cmidrule(l){2-5} 
\multicolumn{1}{c}{}                          & E2E Coverage & E2E Precision & E2E Coverage & E2E Precision \\ \cmidrule(r){1-5}
Dataset 1                                     & 83.23\%      & 72.15\%       & 90.23\%      & 71.40\%       \\
Dataset 2                                     & 95.68\%      & 89.96\%       & 96.28\%      & 90.72\%       \\
Dataset 3                                     & 58.32\%      & 50.61\%       & 55.79\%      & 50.74\%       \\ \midrule
\textbf{Average}                                       & 79.08\%      & 70.91\%       & 80.77\%      & 70.95\%      \\ \bottomrule
\end{tabular}%
}
\caption{E2E metrics for System 1 and System 2}
\label{tab:e2e-metrics}
\end{table}

\begin{table}[]
\centering
\resizebox{\columnwidth}{!}{%
\begin{tabular}{@{}ccccccc@{}}
\toprule
\multirow{2}{*}{} & \multicolumn{2}{c}{\textbf{Relation Prediction}} & \multicolumn{2}{c}{\textbf{Entity Linking}} & \multicolumn{2}{c}{\textbf{Answer Prediction}} \\ \cmidrule(l){2-7} 
                  & \textbf{Coverage}      & \textbf{Precision}      & \textbf{Coverage}    & \textbf{Precision}   & \textbf{Coverage}     & \textbf{Precision}     \\ \cmidrule(r){1-7}
Dataset 1         & 90.65\%                & 90.94\%                 & 93.15\%              & 91.62\%              & 89.63\%               & 88.87\%                \\
Dataset 2         & 98.68\%                & 96.49\%                 & 98.28\%              & 95.65\%              & 97.36\%               & 95.41\%                \\
Dataset 3         & 68.32\%                & 67.92\%                 & 70.84\%              & 70.05\%              & 63.87\%               & 62.60\%                \\ \midrule
\textbf{Average}  & 85.88\%                & 85.12\%                 & 87.42\%              & 85.77\%              & 83.62\%               & 82.29\%                \\ \bottomrule
\end{tabular}%
}
\caption{Component Level Metrics for System 1}
\label{tab:comp-metrics}
\end{table}

\begin{figure}[h!]
\centering
\includegraphics[width=\columnwidth]{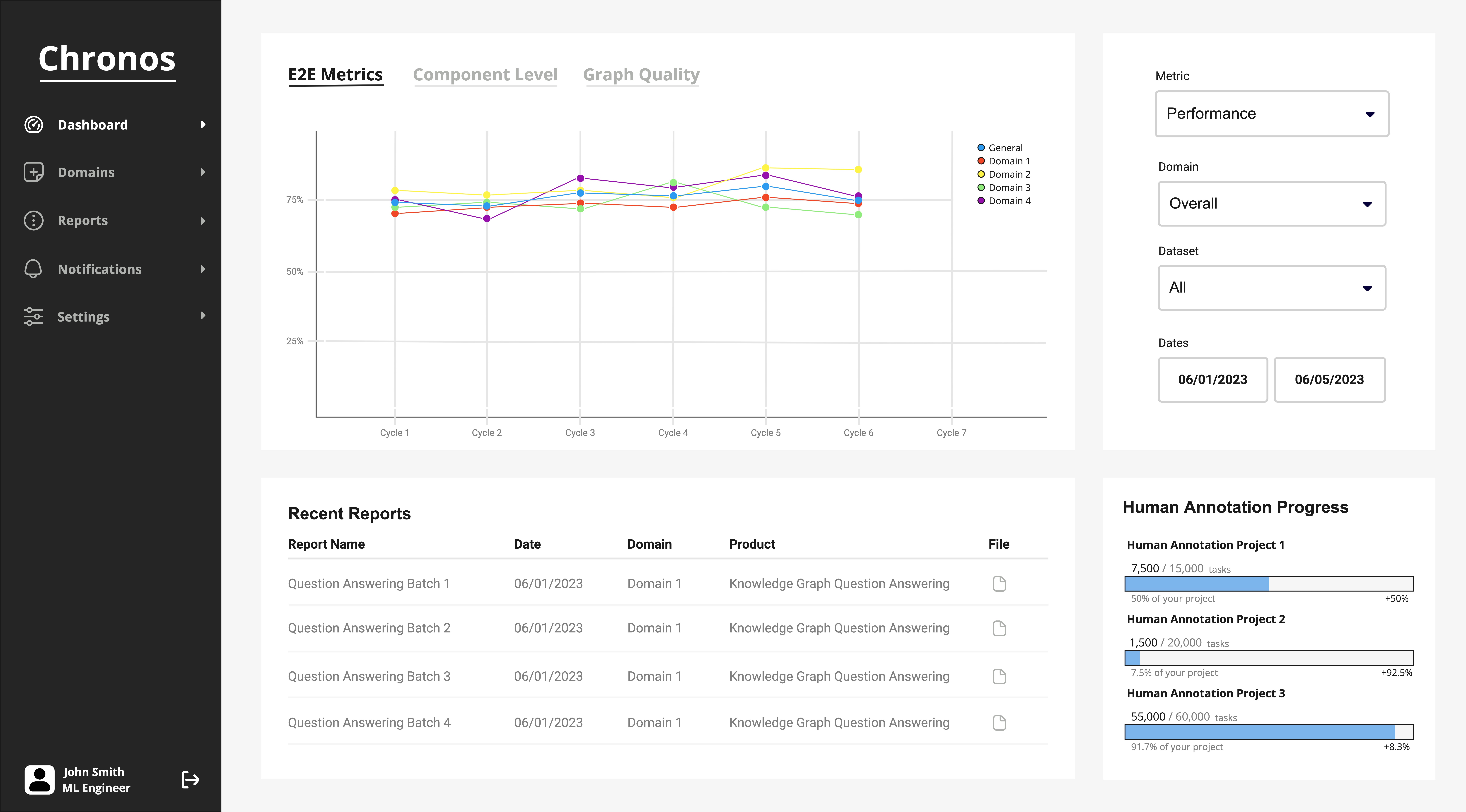} 
\caption{Metrics Dashboard for tracking KGQA and KG quality}
\label{metric-dashboard-fig}
\end{figure}

Based on these metrics, we visualize the loss buckets for KGQA defined in Section \ref{sec:loss_buckets}
in the form of a sankey diagram in the dashboard (Figure \ref{metric-dashboard-fig}).
Some examples for each of the buckets are included in Table \ref{table:loss-bucket}.
\begin{table*}
    \small
    \begin{tabular}[width=\textwidth]{|p{3.5cm} |p{4.5cm} |p{6.5cm} |}
        \hline
            \bf Loss Bucket & \bf Query & \bf Analysis \\
        \hline
        Unsupported Relation & how many cells in human body & Relation needs to be added to retrieve fact from KG\\
        \hline
        Relation Prediction Error & when the nexxt Tourde France & Relation not recognized correctly\\
        \hline
        Entity Prediction Error & Who won Paris & Ambiguous entity - Paris Masters and Paris–Roubaix\\
        \hline
        Missing Entity in KG & Who is princess noor horse & This might be due to entity/alias not present\\
        \hline
        Missing Fact in KG & When is the oscars in 2026 & Fact is absent because it's not published yet\\
        \hline
        \end{tabular}
        \caption{Error analysis on queries from loss-bucketization process}
    \label{table:loss-bucket}
\end{table*}

These loss buckets under the KG error category provide further analysis of the Knowledge Graph. We define three graph quality metrics for this:
\vspace{-6pt}
\begin{itemize}
\item  Accuracy measures the correctness of facts in the graph as observed in the queries for each category. 
\vspace{-6pt}
\item Freshness measures the proportion of queries that are answered incorrectly due to stale information in the graph. Time sensitive relations like net worth, event date, are monitored through this metric. 
\vspace{-6pt}
\item Coverage focuses on evaluating queries that are answered incorrectly due to missing facts. Relevant relations include children, cast members, performer etc. 
\vspace{-6pt}
\end{itemize}

Since part of the data in the graph is time-sensitive, the component-level evaluation needs to run regularly to check if those data are up to date. This requires the capability of continuous evaluation in the pipeline. In the design of the evaluation pipeline, a function to calculate delta facts from previous refresh was added. Delta facts with other time-sensitive facts marked by the human annotators, are extracted and presented in a refresh task for re-assessment each month. This continuous evaluation ensures the freshness of the data in the graph.  
The evaluation of knowledge graph not only measures the component-level quality as presented in the KGQA system, but also provides data driven approach for quality improvements.

The main section of the dashboard in Figure \ref{metric-dashboard-fig} shows accuracy scores for each query set across categories, which facilitates comparison among datasets and domains. Continuous evaluation of the same query set at monthly cadence are shown in a sequence to help monitor changes over time and identify any potential regression. Besides accuracy, top relations for incorrect answers are extracted and presented in a ranked list to help drive decision making for the end-to-end QA system. 

\section{Downstream Impact}
Using the insights provided by the evaluation framework for KGQA, we can help the stakeholders in better decision making. It is evident that the unsupported relations need to be discussed with product management to ensure timely addition to the system. The big problematic areas highlighted are the relation classification and entity linking errors - so we need to invest in improving the algorithms for these two components, after studying the error buckets and their root causes. From our study, a lot of the queries which require 
need to be addressed from a model accuracy perspective by incorporating other signals from user logs for increasing precision.
In addition to that, we would also require a user-experience changes to surface the appropriate answer in minimum number of user interactions. 
The missing entity or fact in KG can help us identification of low-coverage domains and targeted data acquisition.
The overall quality of the graph suggests more investments on fact accuracy and curation on certain domains.

\section{Lessons learned}
Since KGQA is a complex system which is always adapting to user-requirements, Chronos is designed to be adaptable. We found that a modular approach helps in the development and maintenance of Chronos with minimal overhead. Continuous addition of fresh utterances through our data collection process listed in \ref{sec:data_collection} helps us in avoiding undetected regressions.
The quality of human annotations need to be continuously evaluated using feedback from graders and annotation verification by domain-experts on the team. Also, after completing the loss bucket analysis, the source data is reviewed and validated for its quality as a mitigatory step by domain experts. This is to ensure the losses are not derived from the source data, and the integrity of the metrics. 

\section{Conclusion}
In this work, we describe a comprehensive KGQA evaluation system called Chronos, deployed in production. The components of the system help to address crucial challenges around complexity \& generality of KGQA system, and repeatability of the evaluation process. The system considers (1) various data slices for evaluation through the data collection process, (2) human annotation process to retrieve labels at component level, (3) loss-bucketization of errors and (4) detailed metrics tracking through a dashboard - highlighting failures areas of the KGQA system. This helps in decision-making for stakeholders, debugging for developers and continuous monitoring of the service in an industry setting. We hope this framework provides basis for other industry-scale KGQA system.

\section*{Limitations}
The Chronos system
is built under the assumption of typical KGQA architecture described in Section \ref{prelim} and in Figure \ref{sys-eval-fig}. Not all KGQA systems have similar component-level design, and the evaluation approach needs to be
adapted accordingly.
Although this architecture ensure ease of debugging with component level metrics, it will be especially challenging 
to obtain and use the component-level metrics in certain end-to-end KGQA systems. The data collection component, E2E metrics and majority of the human annotation process is still applicable to most KGQA systems.
Also, to ensure a high-quality industry-scale KGQA system in production we require human-in-the-loop for data annotation process as well as domain experts to debug the output of the system. This process is often expensive, hence the use of LLM to reduce the workload for domain-experts can be explored as future work.


\newpage

\bibliography{acl_latex}
\bibstyle{acl_natbib}

\appendix


\end{document}